\documentclass[runningheads]{llncs}

\usepackage{graphicx}
\usepackage{url}
\usepackage{hyperref}
\usepackage{adjustbox}
\usepackage[ruled,vlined]{algorithm2e}

\usepackage[noend]{algorithmic}
\usepackage{amsmath}
\usepackage{amsfonts}
\usepackage{amssymb}
\usepackage{stackengine}

\newcommand\xrowht[2][0]{\addstackgap[.5\dimexpr#2\relax]{\vphantom{#1}}}
\begin{document}

\title{Single Sample Feature Importance: An Interpretable Algorithm for Low-Level Feature Analysis}

\titlerunning{Single Sample Feature Importance}
%
%
\author{Joseph Gatto\inst{1,2} \and
Ravi Lanka\inst{1}\and
Yumi Iwashita \inst{1} \and
Adrian Stoica \inst{1}}
\authorrunning{J. Gatto et al.}
%
\institute{Jet Propulsion Laboratory, California Institute of Technology \and
Columbia University\\}
\maketitle              
\begin{abstract}
Have you ever wondered how your feature space is impacting the prediction of a specific sample in your dataset? In this paper, we introduce Single Sample Feature Importance (SSFI), which is an interpretable feature importance algorithm that allows for the identification of the most important features that contribute to the prediction of a single sample. When a dataset can be learned by a Random Forest classifier or regressor, SSFI shows how the Random Forest's prediction path can be utilized for low-level feature importance calculation. SSFI results in a relative ranking of features, highlighting those with the greatest impact on a data point's prediction. We demonstrate these results both numerically and visually on four  different datasets.

\keywords{Interpretable Machine Learning \and Random Forest \and Feature Importance}
\end{abstract}
\section{Introduction}


The interpretability of a machine learning (ML) algorithm is critical to many data analysis tasks. Perhaps the most popular motivating example involves ML interpretability in medical data analysis. Let us imagine we have a set of patients in our dataset, some with heart disease, some without. Given a set of features about these patients, one may not only be concerned with the accurate prediction of heart disease, but of the features most important to the success of this prediction task. Let us denote these important features as the \textit{feature importances} of our medical dataset. Such feature importances can help medical professionals gather useful insights regarding the prevention of heart disease. 

Currently, one of the most popular methods to achieve such insights is by using the feature importances calculated when training a Random Forest (RF) \cite{Breiman2001}. RF  is an ensemble learning technique which utilizes numerous decision trees for regression and classification tasks. As described in the original paper, RF allows us to quantify feature importance given the interpretable nature of the underlying decision tree structure. 

RF solves the interpretability problem by providing knowledge of feature importances on a global scale. That is, RF inherently answers the question "What features best separate all data points?" Yet, what if one desires to know how important a feature is to the prediction of a specific sample? For example, what if we wish to gather feature importance insights for a specific patient in our medical dataset? Maybe he or she had an outlying characteristic we wish to explore on a \textit{low-level}? This is the type of insight Single Sample Feature Importance (SSFI) aims to provide.

In general, SSFI builds on RF by answering the question, ``What features are contributing most to the prediction of the target variable \textit{for a single sample} in our dataset?" Our SSFI algorithm exploits the existing properties of RF to quantify the contribution of each feature to the prediction of a given sample.



\section{Related Work}

When it comes to analyzing how all samples are impacted by the feature space during prediction, there are a number of existing methodologies. Various forms of Multiple Linear Regression coefficient analysis \cite{standardize_coef} \cite{Dominance_Analysis} \cite{MR_PY} \cite{heuristic_importance} \cite{RIA} \cite{multifaceted} have been utilized to answer questions pertaining to feature importance. Upon the development of ensemble-tree learning algorithms \cite{Breiman2001}, RF has become a popular tool for feature importance calculation. The specifics of RF's default feature importance calculation is discussed in Section \ref{VarImport}. Variations of this approach, such as Permutation Importance \cite{perm_import} combat bias produced by RF in the presence of categorical class imbalance. However, none of the aforementioned approaches allow for investigation of feature importances for a single sample. This is the problem SSFI aims to investigate. 

The notion of using RF for low-level feature analysis was first introduced by \cite{Palczewska2014}. However, SSFI expands on this idea by redefining the way one calculates the importance of the feature at a certain node in a Decision Tree. Our node importance function is detailed in Section 4.1. 

A general solution to the low-level model interpretability problem is LIME \cite{lime}, which can explain the prediction of any classifier or regressor via local approximation of an interpretable model. Both LIME and SSFI provide a relative ranking of feature importances during the prediction of a given sample. However, LIME is built to provide a \textit{general} solution to the interpretability problem, while SSFI should be viewed as an extension of RF. We compare the performance of both the general and RF based solutions within a classification setting in Section 5.2.

When analyzing image data,  \cite{CMAP} shows how Class Activation Maps (CMAP) can be used to identify image regions with the greatest impact on a prediction made by a Convolutional Neural Network. Since SSFI is built upon RF, which is generally unable to learn complex image data, CMAPs can be difficult to compare with SSFI. However, when presented with small grayscale images learnable by RF, we are able to compare the important pixels extracted from both algorithms. We compare the results of SSFI with CMAPs in Section \ref{CMAP_REF}.

\section{Random Forest} \label{RandomForest}
RF is a popular algorithm used for both classification and regression tasks. RF is an ensemble of decision trees, which makes them an attractive predictor for a variety of reasons. First, Decision Trees are non-parametric, allowing for the modeling of very complex relationships without the use of a prior. Furthermore, Decision Trees efficiently utilize both categorical and numeric data, are robust to outliers, and provide an interpretable modeling of the data \cite{1407.7502}. 

More formally, Decision Trees can be understood as recursive partition classifiers. Let us denote a learning set $L=\{(X_1, Y_1), (X_2, Y_2), \dots, (X_n, Y_n)\}$ where each $\{X_1, X_2, \dots, X_n\} \in X$ is a $k \times$1 input vector containing $k$ explanatory variables and each $\{Y_1, Y_2, \dots, Y_J\} \in Y$ is a continuous value (regression) or class value (classification) corresponding to the respective target. In the case where $Y = \{y_1, y_2, \dots y_J\}$, Decision Trees aim to recursively partition the inputs X  into the $J$ subsets which minimizes
\begin{equation}
\label{eqn:G}
G = \sum_{i=1}^J p(i) (1-p(i))
\end{equation}
where $p(i)$ is the probability of picking a sample with class label $i$. $G$, known as the \textit{Gini impurity} \cite{Breiman1983ClassificationAR}, quantifies the quality of a split by how mixed the classes are in the $J$ groups created by the split. A perfect split results in $G = 0$, which means the probably of picking class $i$ in subset $i$ is 1. 

Decision Trees suffer from high variance as a single tree is highly dependent on its given training data. To combat this issue, one can apply a technique called Bootstrap Aggregating (Bagging) \cite{Breiman1996_}. Given our Learning set $L$, bagging creates $B$ Decision Trees which are trained on a random subsample of $L$ (with replacement) of size $n$. Thus, given $B$ Decision Trees, our Bagging-based prediction of our target variable is
\begin{equation}
\label{eqn:yhat}
\hat{Y} = \frac{1}{B}\sum_{i=1}^B f_i(x)
\end{equation}
where $f_i$ is the output of Decision Tree $i$ given some input vector $x$ of size $n$. This approach reduces the inherent Variance problem related to single Decision Tree learners by training many trees on slightly different datasets. 

A limitation of Bagging is that each tree uses the same sampling procedure for each tree, allowing for highly correlated trees which may have equivalent (or very similar) split points. RF solves this by limiting the number of features to be considered by each tree during the bagging procedure. For Classification problems, the number of features used in each split in a dataset with $p$ features is usually $\sqrt{p}$. This results in lower correlation between trees, higher diversity in predictions, and an improvement in overall accuracy. 

\subsection{Feature Importance} \label{VarImport}
The most popular feature importance measure utilizing RF is known as the Gini Importance \cite{Breiman1983ClassificationAR}. The Gini Importance is implicitly calculated during training as it is a product of the Gini Impurity used to calculate Decision Tree splits. That is, when training a Decision Tree, one can calculate how the selection of a feature $\theta$ at node $n$ in tree $t$ contributes to the minimization of the Gini Impurity. We can extend this idea to RF and simply observe a feature's total impact on the Gini Impurity's decrease throughout all trees in a RF.


Let $T$ represent a trained RF consisting of $K$ trees. For each tree, let there be $n_i$ nodes in tree $t_i$, the importance of feature $\theta$ at $n_i$ is computed as:

\begin{equation}
\label{eqn:feat_import}
Importance_{\theta}(n_i) = w_iG_i - w_{left}G_{left} - w_{right}G_{right}
\end{equation}

where $w_i$ is the weighted number of samples (i.e. number of samples for a node divided by total number of samples) that reach $n_i$, $G_i$ is the Gini Impurity of $n_i$, $w_{left}$ and $w_{right}$ represent the weighted number of samples reaching the left child node and right child node of $n_i$ respectively and $G_{left}$ and $G_{right}$ represent the Gini Impurity of the left and right child nodes respectively. Thus the importance of $n_i$ represents the Gini decrease provided by the feature used to split at node $n_i$. 

Perhaps more intuitively, it might be useful to think about the Gini decrease as:
\begin{equation}
\label{eqn:import_}
\nabla G(t) = G(t)_{\mathrm{parent}} - G(t)_{\mathrm{split\:left}} - G(t)_{\mathrm{split \: right}}
\end{equation}
as the importance of a node is quantified by how much it contributes to the minimization of the Gini Impurity. Thus, given some feature $\theta$ used to split node $n_i$, the larger $\nabla G(t)$ is, the greater impact $\theta$ has on Impurity minimization.

One can thus calculate the total Gini Importance for variable $\theta$ as:
\begin{equation}
\label{eqn:import2}
I_G(\theta) = \sum_t^T \sum_n^N \nabla G_\theta (n,t)
\end{equation}
Which is the total decrease in impurity across all nodes $n$ and all trees $t$ in a forest $T$ contributed by feature $\theta$. Note that Eq. \ref{eqn:import2} assumes that $\nabla G_\theta$ at a node $n$ for which feature $\theta$ is not present produces 0 change in Gini decrease. 

\section{Single Sample Feature Importance}
A limitation of the variable importance calculation as described in section \ref{VarImport} is that it only provides insight into how important a feature is in the context of global Gini Impurity minimization. What if one desires to know how important a feature is to the prediction of a specific sample? In this section, we detail the SSFI algorithm and outline all underlying assumptions used during our calculation. 

\subsection{Algorithm}
\begin{algorithm}[H]
\SetAlgoLined
\caption{SSFI}
\label{alg:alg}
\begin{algorithmic}
    \STATE SSFI Results $ = $ \{\}
    \FOR{sample \textit{in} Dataset} 
        \STATE Feature Importances = \{\}
        \STATE Train Data = Dataset - sample
        \STATE Test Sample = sample
        \STATE RF\_Model $= \mathrm{RF}.train(\mathrm{Train \: Data})$
        \WHILE{RF\_Model.$predict$(Test Sample)} 
            \FOR{tree \textit{in} RF\_Model}
                \STATE Predict Path = tree.$predict$(Test Sample)
                \FOR{ (node, feature name) \textit{in} Predict Path}
                    \STATE $ \mathrm{Feature \: Importance}  = node\_importance(\mathrm{node})$
                    \STATE Feature Importances[feature name] += Feature Importance
                \ENDFOR 
            \ENDFOR 
        \ENDWHILE
        \STATE SSFI Results[sample] = Feature Importances\\
    \ENDFOR
\end{algorithmic}
\end{algorithm}


\noindent Consider a trained RF model $T$ with $k$ trees, where each tree $t_i \in T$ has depth $d_i$, and $n_i$ total nodes. Let $P_i^{\phi} = \{p_1, p_2, \dots, p_M \}$ be the sequence of nodes used during the prediction of sample $\phi$ given $t_i$. For all nodes $p_m \in P_i^{\phi}$ there exists a split value $s$ which has been calculated using the methods described in Section \ref{RandomForest}. We predict the output of a test sample $\phi = \{\theta_1, \theta_2, \dots, \theta_j \}$ which has $j$ features, each with some feature value $f$. SSFI defines the importance of feature $\theta$ at node $p_m$ as:

\begin{equation}
\label{eqn:ssfi_calc}
I_{\theta}^{p_m}(f,d,s,\alpha) = \frac{1}{1 + \alpha^{-d}} \times |s-f|  
\end{equation}
Where $f$ is the feature value of $\theta \in \phi$ which corresponds to the feature used to split $p_m$, $d$ is the depth of $p_m$, $s$ is the computed split value of node $p_m$, and $\alpha$ is a free tuning parameter. In our experiments, we found $\alpha = 0.9$ to be the optimal value. 

The overall importance of $\theta$ to the prediction of $\phi$ is the sum of all $I_{\theta}^{p_m}$ for all $p_m \in P_i^{\phi}$, and all $t \in T$:
\begin{equation}
\label{eqn:total_ssfi}
SSFI({\theta}) = \sum_i^k \sum_{p_m}^{P^{\phi}_i} I^{p_m}(\theta)
\end{equation}
Note that Eq. \ref{eqn:total_ssfi} assumes a contribution of 0 by feature $\theta$ at node $p_m$ if $\theta$ does not correspond to the node used to split $p_m$. Eq. \ref{eqn:ssfi_calc} highlights how SSFI quantifies importance. The first component $\frac{1}{1 + \alpha^{-d}}$ implies that a feature which occurs earlier in the tree (i.e. closer to the root) will have a greater impact on the prediction. This follows one's intuition as the first split in a Decision Tree influences the longest sequence of split decisions in the tree. The later component $|s-f|$ can be understood as the ``confidence" of a split, where ``confidence" simply means the distance of the feature value from the split value. All together, Eq. \ref{eqn:ssfi_calc} rewards features that were used early on in prediction and contain large distances from computed node split values. Eq. \ref{eqn:total_ssfi} has a similar interpretation to Eq. \ref{eqn:import2}, as we simply sum these importance values across all predictions made by a RF. 
\subsection{Assumptions}
It is important to note the assumptions that need be made when utilizing SSFI. First, we assume the samples in our learning set and our test sample come from the same dataset. The trained RF model used during calculation has never seen the SSFI sample being analyzed. Yet, since we derive SSFI from this trained model, the test sample must share the underlying structure of the training dataset. In practice, we generate SSFI for all samples by performing Leave-One-Out (LOO) cross validation \cite{ref1} where given some learning set $L=\{(X_1, Y_1), (X_2, Y_2), \dots, (X_n, Y_n)\}$ where each sample is from the same distribution, one can iteratively calculate SSFI for the held out sample while training the RF model on the other $n-1$ remaining samples. This is displayed in Algorithm \ref{alg:alg}.

Furthermore, it is important for SSFI that the trained RF model $T$ is able to partition the data well and produce accurate predictions. Proper testing of RF's ability to separate a given dataset must be performed prior to SSFI's utilization. If utilizing LOO cross validation, one may verify the viability of SSFI by analyzing the error during LOO and ensuring the model is able to predict test samples accurately. 

\section{Results}
Assessing the performance of SSFI is non-trivial as one generally cannot know the ground truth single-sample  feature importances for a given dataset. Thus, we have contrived a series of experiments that aim to test SSFI's validity. In this section, we first show how SSFI as a method of feature selection compares to both LIME and the traditional Gini Importance measure. Later, we perform a more qualitative analysis by examining the pixels SSFI deems most important during image classification. 

\subsection{Evaluation Metrics}
To evaluate model performance on experiments with numeric data, we use the Coefficient of Determination ($r^2$):

\begin{equation}
r^2 = \left(\frac{{}\sum_{i=1}^{n} (f_i - \overline{f})(y_i - \overline{y})}
{\sqrt{\sum_{i=1}^{n} (f_i - \overline{f})^2(y_i - \overline{y})^2}}\right)^2
\end{equation}

Where $n$ is the number of samples in our test set, $f_i$ is the predicted value of ground truth value $y_i$. $\overline{f}$ and $\overline{y}$ represent the mean of the predicted values and ground truth values respectively. 

For image classification, we quantify accuracy  as:

\begin{equation}
    \mathrm{accuracy} = \frac{1}{n} \sum_{i=0}^{n-1} 1(\hat{y}_i = y_i)
\end{equation}

Where $n$ is the number of samples in our test set and $\hat{y}_i$ is our predicted class for sample $n_i$ which has ground truth label $y_i$.

\subsection{Feature Selection using SSFI} 

To begin our evaluation, we pose the question: How do SSFI features compare to the top features calculated by RF and LIME? To investigate, let us construct four  different experiments: 

\begin{enumerate}
    \item Evaluate model performance using Leave-One-Out (LOO) cross validation, where at each iteration of LOO, the feature space used to predict the target variable dynamically adjusts to utilize the pre-computed SSFI features for the test sample. 
    \item Evaluate model performance using LOO cross validation, where at each iteration of LOO, the feature space used to predict the target variable dynamically adjusts to utilize the pre-computed LIME features for the test sample. 
    \item Evaluate model performance using LOO where, throughout the cross validation process, the feature space remains static. The static feature set is chosen by a RF model previously trained on the same dataset. 
    \item Evaluate model performance using LOO and a random feature set. This experiment will establish a baseline result for comparison with 1,2, and 3. 
\end{enumerate}

\noindent Each experiment is run 50 times to account for slight variations in performance due to the random component of RF models. Thus, all results are the average $R^2$ score after 50 experiments. Furthermore, each model will only be trained on the top-3 features produced by each selection method. We evaluate feature performance using both linear (Linear Regression \cite{regresh}) and non-linear (RF) models. All RF models used for prediction are fit using 100 estimators and all model inputs are normalized between $[0,1]$. 

\begin{table}[]

\caption{Results on the Wine Classification Data Set \cite{Dua:2019} which is a 3-class dataset with 178 samples and 13 features.}
\begin{tabular}{|c|c|c|}

\hline \xrowht[()]{6pt} 
\textbf{Feature Selection Method} & \textbf{$R^2$ Random Forest} & \textbf{$R^2$ Linear Regression} \\\hline 
SSFI & 92.0\% &  80.6\% \\ \hline
LIME & 84.8\% & 74.5\% \\ \hline 
Random Forest & 66.9 \% & 62.6\%  \\ \hline
Random & 57.8 \% &  52.5\% \\ \hline
\end{tabular}
\label{tab:table1}
\end{table}

\noindent The results in Table \ref{tab:table1} highlight how the dynamic SSFI feature set selects features that outperform both LIME and RF. By validating performance on both RF and Linear Regression, we ensure our features generalize outside of the presence of RF, with which SSFI is generated by. In general, this suggests SSFI is identifying important sample-level information about the feature space not captured by the other approaches. 

\begin{table}[]
\caption{Results on the Breast Cancer Data Set \cite{Dua:2019} which is a binary classification dataset with 669 samples and 30 features.}
\begin{tabular}{|c|c|c|}
\hline \xrowht[()]{6pt} 
\textbf{Feature Selection Method} & \textbf{$R^2$ Random Forest} & \textbf{$R^2$ Linear Regression} \\\hline 
SSFI & 80.1\% &  74.6\% \\ \hline
LIME & 79.7\% & 75.4\% \\ \hline
Random Forest &  46.6\% & 48.9\%  \\ \hline
Random &  39.5\% &  53.2\% \\ \hline
\end{tabular}
\label{tab:table2}
\end{table}

The results in Table \ref{tab:table2} were conducted under the same conditions as Table \ref{tab:table1}, only with the Breast Cancer Dataset \cite{Dua:2019}. Once again, we find that SSFI calculated features vastly outperform the static feature set produced by the Random Forest model. However, in this setting, LIME and SSFI generally select overlapping feature sets that produce very similar results. It may be the case that LIME performs better on binary classification tasks, but nonetheless shows SSFI and LIME are at least in agreement. 

These experiments highlight the existence and accessibility of low-level feature information. Furthermore, SSFI is clearly not guessing as the dynamic feature space significantly outperforms both globally important features and randomly selected features. Thus, SSFI features have been shown to be reliable predictors for their given test samples when presented with a dataset learnable by RF. However, we must note that we do not wish to compare the overall quality of LIME and SSFI.  LIME's intention is to provide a general solution to the low-level interpretability problem. Furthermore, LIME provides more information than SSFI, such as what features caused a certain class \textit{not} to be predicted. Thus, the SSFI-LIME comparison simply serves to help provide a quantitative analysis in the absence of ground-truth labels. Nonetheless, the Wine classification results may suggest that, in the situations where RF can separate a dataset extremely well, RF-based interpretability methods may allow for greater knowledge extraction. 

\subsection{Visual Analysis} \label{CMAP_REF}
We may also interpret the SSFI feature selection choices visually by training a RF classifier on image data. RF is generally not a viable choice for image classification, but in the presence of simple grayscale images, can produce accurate results. For example, when training a RF classifier on the MNIST Handwritten Digit dataset \cite{lecun-mnisthandwrittendigit-2010}, we achieve $92\%$ accuracy. When we run SSFI on this trained classifier, we can visualize the most important pixels used during classification. 

\begin{figure}[!h]
\centering
    \includegraphics[scale =.5]{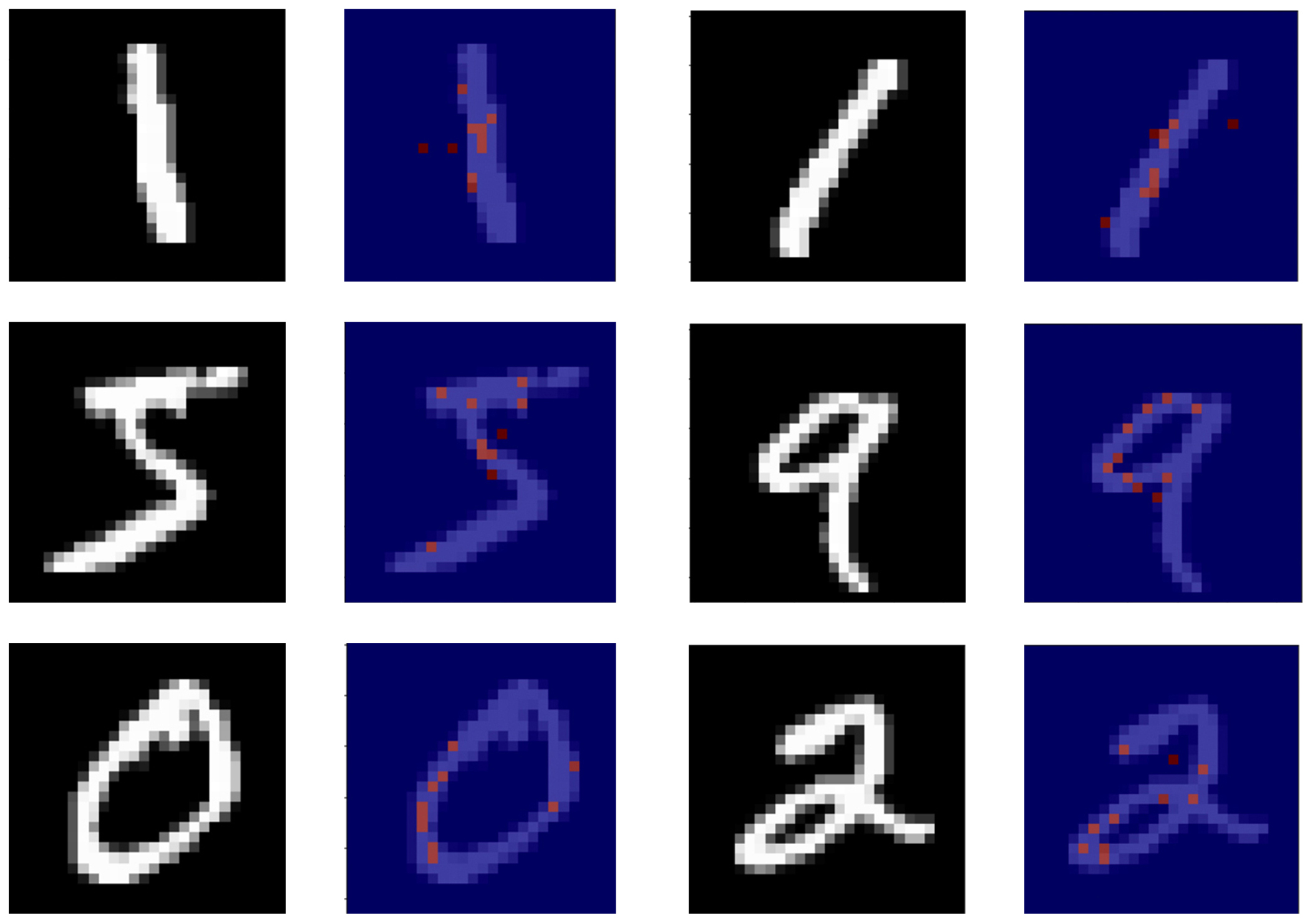}
    \caption{Ground truth MNIST images (left) vs. the top-10 most important SSFI-extracted pixels highlighted in red (right).}
    \label{fig:mnist_results}
\end{figure}

Figure \ref{fig:mnist_results} shows SSFI's effectiveness in extracting useful features when predicting top MNIST features. We find that SSFI consistently identifies pixels that construct each digit during it's feature extraction. 

We perform a second level of visual analysis with the Fashion MNIST dataset \cite{xiao2017/online} which contains 60,000 grayscale clothing images. This time, we compare SSFI with the Class Activation Map (CMAP) \cite{CMAP}, which is a deep learning strategy that visualizes the Global Average Pooling layers of a trained Convolutional Neural Network (CNN) to highlight the regions of an image which are important for the classification of that image. To do so, we use each 28$\times$28 in Fashion MNIST to train a ResNet16 \cite{resnet} CNN which can classify Fashion MNIST with 95\% accuracy. Next, we train a RF classifier on Fashion MNIST which obtains 85\% accuracy.  Given our trained CNN, we visualize the CMAP for a sample of images and compare these regions to the pixels extracted by SSFI of the same sample.

\begin{figure}[!h] 
\centering
    \includegraphics[scale =.45]{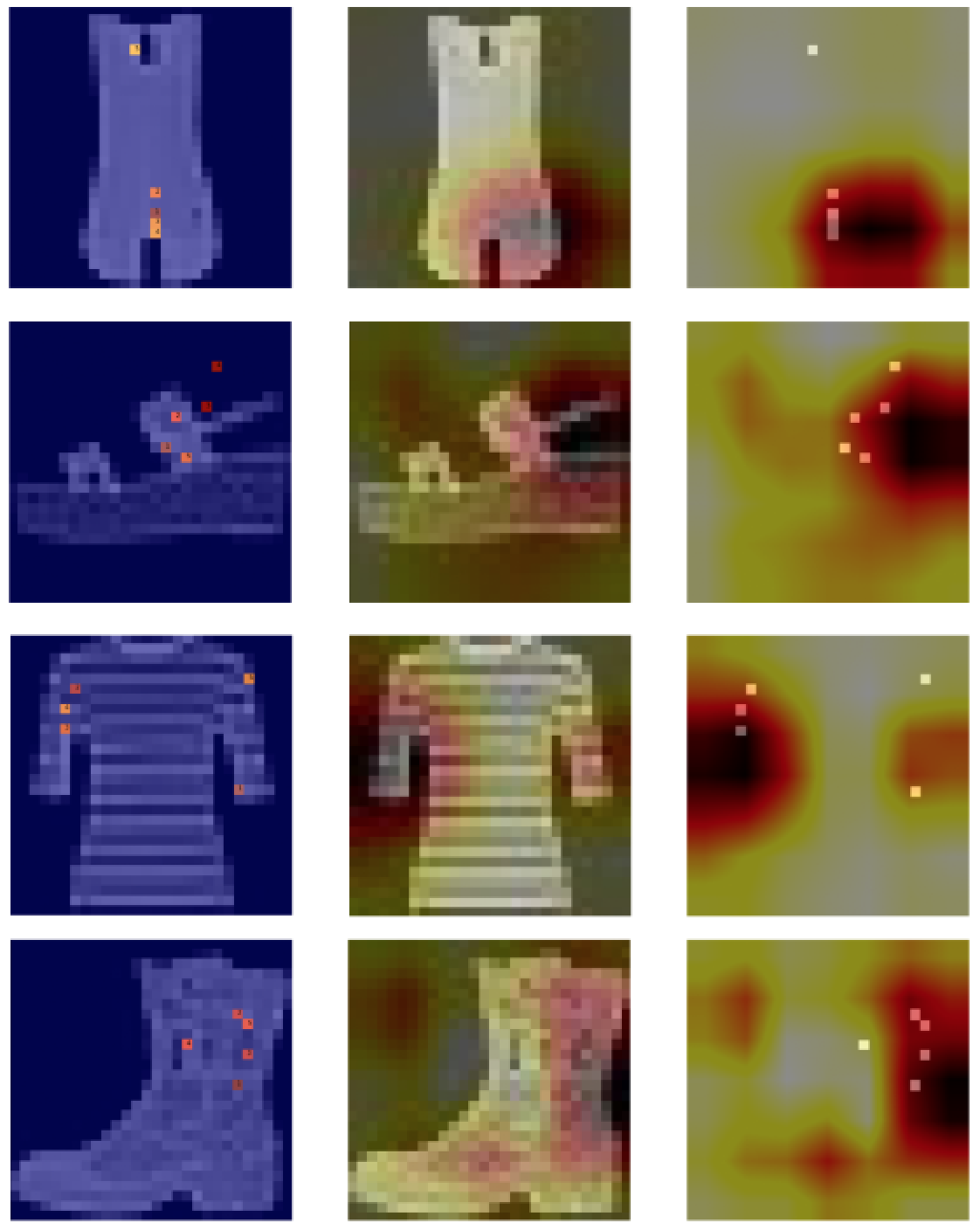}
    \caption{Comparison of SSFI top 5 pixels (Left) vs important CMAP regions colored in red (Center). The rightmost column highlights SSFI extracted pixels that fall into the CMAP regions where red regions are the most important CMAP regions. }
    \label{fig:Fmnist_results}
\end{figure}

Figure \ref{fig:Fmnist_results} highlights how the SSFI extracted pixels generally fall in the extracted CMAP regions. The displayed results are what we deemed a representative sample of our findings. For example, long thing tops, as shown in row 1 of Figure \ref{fig:Fmnist_results}, often had a dense region near a corner of the clothing deemed most important. Furthermore, rows 2 and 4 highlight foot ware, where both SSFI and CMAP almost always found the back of the shoe/boot to be important. Finally, row 3 shows the prediction of a shirt, where the sleeves were often important for classification. These results show that when RF is able to accurately learn how to separate image data, it is in agreement with the CNN regarding pixels important for classification. 

In an attempt to quantify the SSFI and CMAP comparison, we also calculated how often SSFI pixels were appearing in CMAP regions. To do so, we define an evaluation metric which returns 1 if any of the top 5 pixels from SSFI appear in the red "most important" CMAP regions, and 0 otherwise. This experiment assumes the CMAP to be the ground truth, and is performed only to explore if SSFI agrees with the deep learning solution to this problem. In our experiment, we use 1000 random samples from Fashion MNIST. However, only trials where both the RF used to predict the test sample, and the CNN used to generate the CMAP make a correct prediction are considered in this analysis. This constraint resulted in only 647 data points being used. 

\begin{table}[!h]
\caption{SSFI vs CMAP results, where each column denotes how often SSFI pixels occured in CMAP important regions for a given class in the Fashion MNIST dataset. Note that the Bag class is not included due to limited number of samples.}
\begin{adjustbox}{width=\columnwidth,center}
\begin{tabular}{|c|c|c|c|c|c|c|c|c|c|c|}
\hline
 & \textbf{T-shirt/top} & \textbf{Trouser} & \textbf{Pullover} & \textbf{Dress} & \textbf{Coat} & \textbf{Sandal} & \textbf{Shirt} & \textbf{Sneaker} & \textbf{Ankle Boot} & \textbf{Total} \\ \hline
\textbf{Number of Samples} & 86 & 86 & 52 & 74 & 50 & 71 & 55 & 100 & 73 & 647 \\ \hline
\textbf{Accuracy} & .755 & .906 & .731 & .459 & .621 & .915 & .623 & .780 & .876 & .736 \\ \hline
\end{tabular}
\end{adjustbox}
\label{tab:hey}
\end{table}

Table \ref{tab:hey} displays the results of the SSFI vs CMAP comparison. In general, we find that both algorithms consider similar portions of the region to be most important to classification 73\% of the time. Certain classes, such as Dress, Coat, and Shirt brought about disagreement between algorithms. However, both algorithms are strongly aligned in their region importance for Trouser, Sandal, and Ankle Boot. 

By comparing SSFI with CMAP, we may only conclude that, when RF is able to learn a dataset, it tends to identify the same regions as being important to classification as the corresponding deep learning solution. In general, our qualitative visual analysis has served to highlight the validity of the SSFI algorithm in the absence of ground-truth single-sample feature importances. 

\section{Discussion}
The low-level quantification of feature importance is highly desirable in practices where data samples require individualized inspection.  The conditions for SSFI's success are of course stringent. The requirement of data to be well understood by RF reduces generality but provides a tool for analysis when used in the correct setting. 

This brings about interesting questions for future work. Might we validate SSFI's ability to identify the root cause of anomaly detection problems or perhaps the improvement of classification using a dynamic feature space? In this study, our approach looked to verify the quality of the individualized features extracted by SSFI in the general sense, but future work should investigate the use of single sample importances as tools for solving more specific machine learning problems.

\section{Acknowledgement}
The research was carried out at the Jet Propulsion Laboratory, California Institute of Technology, under a contract with the National Aeronautics and Space Administration. The work of Joseph Gatto was sponsored by JPL Summer Internship Program and the National Aeronautics and Space Administration.

\bibliographystyle{splncs04}
\bibliography{bib}

\begin{thebibliography}{10}
\providecommand{\url}[1]{\texttt{#1}}
\providecommand{\urlprefix}{URL }
\providecommand{\doi}[1]{https://doi.org/#1}

\bibitem{perm_import}
Altmann, A., Toloşi, L., Sander, O., Lengauer, T.: {Permutation importance: a
  corrected feature importance measure}. Bioinformatics  \textbf{26}(10),
  1340--1347 (04 2010). \doi{10.1093/bioinformatics/btq134},
  \url{https://doi.org/10.1093/bioinformatics/btq134}

\bibitem{Breiman1996_}
Breiman, L.: Bagging predictors. Machine Learning  \textbf{24}(2),  123--140
  (Aug 1996). \doi{10.1007/BF00058655},
  \url{https://doi.org/10.1007/BF00058655}

\bibitem{Breiman2001}
Breiman, L.: Random forests. Machine Learning  \textbf{45}(1),  5--32 (Oct
  2001). \doi{10.1023/A:1010933404324}

\bibitem{Breiman1983ClassificationAR}
Breiman, L., Friedman, J.H., Olshen, R.A., Stone, C.J.: Classification and
  regression trees (1983)

\bibitem{standardize_coef}
Bring, J.: How to standardize regression coefficients. The American
  Statistician  \textbf{48}(3),  209--213 (1994),
  \url{http://www.jstor.org/stable/2684719}

\bibitem{MR_PY}
Budescu, D.V.: Multiple regression in psychological research and practice.
  Psychological Bulletin  \textbf{69}(3),  161--182 (1968),
  \url{https://doi.org/10.1037/h0025471}

\bibitem{Dominance_Analysis}
Darlington, R.B.: Dominance analysis: A new approach to the problem of relative
  importance of predictors in multiple regression. Psychological Bulletin
  \textbf{114}(3),  542--551 (1993),
  \url{https://doi.org/10.1037/0033-2909.114.3.542}

\bibitem{Dua:2019}
Dua, D., Graff, C.: {UCI} machine learning repository (2017),
  \url{http://archive.ics.uci.edu/ml}

\bibitem{regresh}
Galton, F.: Regression towards mediocrity in hereditary stature. The Journal of
  the Anthropological Institute of Great Britain and Ireland  \textbf{15},
  246--263 (1886), \url{http://www.jstor.org/stable/2841583}

\bibitem{resnet}
He, K., Zhang, X., Ren, S., Sun, J.: Deep residual learning for image
  recognition (2015)

\bibitem{heuristic_importance}
Johnson, J.: A heuristic method for estimating the relative weight of predictor
  variables in multiple regression. Multivariate Behavioral Research
  \textbf{35},  1--19 (01 2000). \doi{10.1207/S15327906MBR3501\_1}

\bibitem{multifaceted}
Kraha, A., Turner, H., Nimon, K., Zientek, L., Henson, R.: The multifaceted
  concept of predictor importance: Tools to support interpreting multiple
  regression. Frontiers in Psychology  \textbf{3},  1--16 (01 2012)

\bibitem{lecun-mnisthandwrittendigit-2010}
LeCun, Y., Cortes, C.: {MNIST} handwritten digit database  (2010),
  \url{http://yann.lecun.com/exdb/mnist/}

\bibitem{1407.7502}
Louppe, G.: Understanding random forests: From theory to practice (2014)

\bibitem{Palczewska2014}
Palczewska, A., Palczewski, J., Marchese~Robinson, R., Neagu, D.: Interpreting
  Random Forest Classification Models Using a Feature Contribution Method, pp.
  193--218. Springer International Publishing, Cham (2014).
  \doi{10.1007/978\-3\-319\-04717-1\_9},
  \url{https://doi.org/10.1007/978-3-319-04717-1\_9}

\bibitem{lime}
Ribeiro, M.T., Singh, S., Guestrin, C.: "why should {I} trust you?": Explaining
  the predictions of any classifier. In: Proceedings of the 22nd {ACM} {SIGKDD}
  International Conference on Knowledge Discovery and Data Mining, San
  Francisco, CA, USA, August 13-17, 2016. pp. 1135--1144 (2016)

\bibitem{ref1}
Sammut, C., Webb, G.I. (eds.): Leave-One-Out Cross-Validation, pp. 600--601.
  Springer US, Boston, MA (2010). \doi{10.1007/978-0-387-30164-8\_469},
  \url{https://doi.org/10.1007/978-0-387-30164-8\_469}

\bibitem{RIA}
Tonidandel, S., LeBreton, J.: Relative importance analysis: A useful supplement
  to regression analysis. Journal of Business and Psychology  \textbf{26},
  ~1--9 (01 2011). \doi{10.1007/s10869-010-9204-3}

\bibitem{xiao2017/online}
Xiao, H., Rasul, K., Vollgraf, R.: Fashion-mnist: a novel image dataset for
  benchmarking machine learning algorithms  (2017)

\bibitem{CMAP}
Zhou, B., Khosla, A., Lapedriza, A., Oliva, A., Torralba, A.: Learning deep
  features for discriminative localization (2015)

\end{thebibliography}





\end{document}